# A Scalable Machine Learning Pipeline for Building Footprint Detection in Historical Maps

**Annemarie McCarthy**

Munster Technological University

**Abstract**

Historical maps offer a valuable lens through which to study past landscapes and settlement patterns. While prior research has leveraged machine learning based techniques to extract building footprints from historical maps, such approaches have largely focused on urban areas and tend to be computationally intensive. This presents a challenge for research questions requiring analysis across extensive rural regions, such as verifying historical census data or locating abandoned settlements. In this paper, this limitation is addressed by proposing a scalable and efficient pipeline tailored to rural maps with sparse building distributions. The method described employs a hierarchical machine learning based approach: convolutional neural network (CNN) classifiers are first used to progressively filter out map sections unlikely to contain buildings, significantly reducing the area requiring detailed analysis. The remaining high probability sections are then processed using CNN segmentation algorithms to extract building features. The pipeline is validated using test sections from the Ordnance Survey Ireland historical 25 inch map series and 6 inch map series, demonstrating both high performance and improved efficiency compared to conventional segmentation-only approaches. Application of the technique to both map series, covering the same geographic region, highlights its potential for historical and archaeological discovery. Notably, the pipeline identified a settlement of approximately 22 buildings in Tully, Co. Galway, present in the 6 inch map, produced in 1839, but absent from the 25 inch map, produced in 1899, suggesting it may have been abandoned during the Great Famine period.

**Introduction**

Historical maps are a valuable resource, offering detailed information on land use (such as woodland and wetland), transportation networks (including railways and roads), geographical features (like rivers and mountains), and patterns of human settlement, including building footprints. Numerous past studies have focused on the automated extraction of information from historical maps. The extraction of building footprints is of particular interest to many researchers, as it allows for the reconstruction of historical landscapes, and the identification of sites of historical and archaeological interest. [1-6]

In this work, historical Irish maps are examined and used to develop a tailored pipeline which can detect building footprints efficiently, in rural areas with low building density. A classifier-guided, progressively focused segmentation pipeline for building detection in historical maps, is proposed. The method first divides the input map into large tiles, then filters tiles using a lightweight classifier trained to identify "interesting" regions — that is, tiles likely to contain buildings. Tiles identified as featureless or irrelevant are discarded early. This process is repeated recursively, shrinking tile size. A segmentation model is finally applied to the reduced size tiles, to extract building footprints. This strategy reduces unnecessary computation in empty or low-information areas, enhancing overall processing efficiency while maintaining good accuracy.

**Background and Related Research**

Automated methods have been used in previous work to extract a wide range of features from historical and topographic maps, including trees [7], wetlands [8], and buildings [1 – 6]. Traditional approaches include the use of edge detection, texture/colour analysis, geometric approaches and pattern recognition [9 – 11]. These methods still find uses in more modern studies [4, 12], but have increasingly been combined with or replaced by machine learning based approaches, which offer improved generalization and robustness to stylistic variation on maps and aging artifacts such as degraded ink and paper marks, among other advantages. CNN techniques have been used in a diverse range of historical map applications. Smith et al. [7] employed instance segmentation techniques using CNNs to accurately detect, separate and count individual tree symbols on urban historical maps, while O Hara et al. [8] used CNNs to detect wetland symbols on the map series studied in this paper.

Recent studies have addressed the specific problem of detecting building footprints in historical maps using deep learning and in particular, semantic segmentation. Goderle et al. [1, 5] used a CNN based on a resnet101 backbone with DeepLabv3 head to locate building footprints on historical maps from the Franciscan Cadastre. Heitzler et al. [2] employed a similar approach, but utilised an ensemble of 10 FCNNs (fully convolutional neural networks) with unet architecture, to perform precise segmentation of tiled map sheets from the Swiss Siegfried map.

While these approaches demonstrate effective segmentation performance, they rely on uniform tiling and/or full-sheet processing, applying segmentation to every part of the image regardless of relevance. While segmentation has been shown to function well in identifying buildings on historical maps, it is fundamentally a computationally expensive pixel by pixel process [13], with inference times of up to 0.26 seconds per 256 × 256 pixel patch reported or estimated for models such as GMEDN and ARC-Net. [14]

While full map segmentation is required when processing urban maps (as the studies described above primarily address), this approach is poorly suited to maps where structures are sparsely distributed, and analysis of large areas is required. Historical maps of Ireland, for example, would comprise approximately 28 million such tiles, required approximately 2000 hours of processing time, excluding the time required for pre and post processing tasks, and model training. This is likely to present a significant practical barrier to many historians and geographers wishing to utilise information from historical maps.

Various techniques which address the accuracy and efficiency of building footprint detection have been proposed. Zhao et al. [15] addressed the issue of the cost of manual labelling for training purposes, training a deep object attention network to extract building footprints, with only limited training samples. Uhl et al. [16] acknowledged the difficulties posed in processing large map collections, proposing some preliminary analysis techniques which could be applied to assist with processing. In a later study [6], the same authors developed an efficient weakly supervised model, utilizing contemporary geospatial data to generate weak labels for training samples extracted from historical topographic maps, and applying a CNN-based patch classifier in a sliding window manner to perform pixel-level segmentation. This allowed for automated estimation of historical settlement patterns, without manually labelled training data. Despite the inherent uncertainty in the segmentation, recall scores as high as 0.97, and F scores of 0.79 were recorded on some individual test maps. Chen et al. [17] proposed an alternative approach to segmentation for building footprint detection, training a bi-directional cascade network to perform edge detection and processing the results using mathematical morphology techniques. Gobbi et al. [12] combined machine learning and other techniques, presenting a pipeline which first segments the entire map into regions using traditional methods of region-growing and connected component analysis and then classifies land use (e.g. forest, building) using a machine learning model.

Despite significant advances in the automated extraction of features from historical maps, the accurate and efficient large-scale analysis of predominantly rural landscapes remains an underexplored area. Existing approaches, while effective in urban or densely featured regions, become computationally burdensome and inefficient when applied uniformly across sparsely populated rural areas. To address this gap, in this work, a novel classifier-guided, progressively focused segmentation pipeline specifically tailored for historical maps with low building density, is introduced. By filtering out non-informative regions early and applying segmentation only where relevant, this approach significantly reduces computational overhead while preserving detection accuracy. This method offers a scalable solution for the detailed reconstruction of rural built environments.

**Data**

Map sheets have been obtained from the Ordnance Survey Ireland (OSI) historical maps, 6 inch series [18] and 25 inch series [19]. The 6 inch series was compiled between 1829 and 1841 and has a scale of 6 inches to 1 mile. The 25 inch series was compiled between 1897 and 1913 and has a scale of 25 inches to 1 mile. Both are available for viewing through the Irish Townland and Historical Map Viewer [20]. Each map was assembled by individual cartographers, referencing a common set of symbols [21], although individual styles are still evident throughout the map series, especially in the representation of features such as woodland and marsh.

**Methods**

*Fastai Library*

The *fastai* library [22, 23] was developed in 2018 by Jeremy Howard and Sylvain Gugger, with the goal of making deep learning more accessible and easier to use for practitioners and researchers. It runs on a PyTorch [23] backbone. As many of the applications of historical map analysis are found in the humanities (e.g. geography and history), in developing this pipeline, it was desirable that the methods described should be accessible to people with limited or no training in the field of machine learning. While the development of new models in *fastai* necessarily requires knowledge of the field, the application of an existing model requires only basic Python skills, with models deployable using only a .pkl file and any associated classes and functions custom written for the specific model. The use of *fastai* tools should also make the training of comparable models on historical maps from other sources, more accessible to non-specialists.

*Classification Models*

In aiming to detect building footprints across large areas of historical map sheets, twin primary goals exist, those of accuracy and efficiency. To reduce the map areas subjected to computationally expensive segmentation models, the area for analysis is iteratively processed through a series of *n* classifier models, beginning first with a tile size of *a* x *b* pixels. The general process is depicted in Fig. 1. The goal is to progressively remove areas of map which contain no features of interest i.e. buildings in the case of this work.

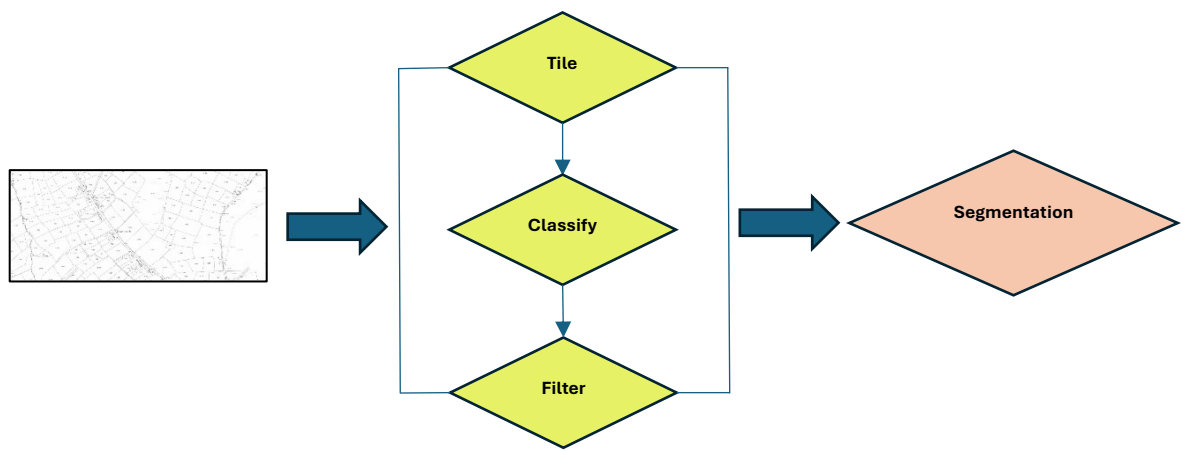

**Figure 1: Workflow model: Map area is iteratively tiled, tiles are classified as 'buildings' or 'no buildings' and filtered on this basis. In this manner, map areas with no buildings present are successively removed from the map. Segmentation is performed only on those areas with buildings present, hence reducing the overall inference time required.**

For this method to succeed in improving the overall efficiency of map processing, it is a prerequisite that the proportion of areas classified as containing buildings must lie below some limit, $R$. This limit is related to the difference between computational times required for segmentation and classification of images. Estimating the exact gain in computational gain is highly model and platform dependent due to issues such as GPU choice, fixed computational overheads, memory limitations, model architecture, choices regarding image batching. Hence, it is difficult to establish this limit absolutely, for all scenarios. However, a sense of how it behaves can be obtained by considering the computational time required in the following simplified case.

Let $t_s$ equal the time required, per pixel, for a segmentation operation, and $t_c$ represent the time required, per pixel, for a classification operation. For the $n = 0$ case, the image is segmented in its entirety, with no regions omitted. The time required per pixel (for map processing) then is simply $t_s$. For the $n = 1$ case, a single classifier is first run on the tiled image and this is followed by a segmentation operation on the tiles of interest. In this case, the time required per pixel is $t_c + R t_s$.

It follows, that for any $n$, the computational time required per pixel, $T$, can be expressed as the sum of a geometric series plus a scaled term.

$$T = \sum_{i=0}^{n-1} R^i t_c + R^n t_s$$

Assuming $t_s$ and $t_c$ can be related by the equation. $t_s = A t_c$, then for a computational gain to be made by using classifiers, the following condition must be satisfied.

$$R < 1 - \frac{1}{A}$$

In the work presented here, the values of $R$ and $A$ are approximately 0.4 and 5, hence satisfying this limit. In maps with a greater density of buildings (e.g. urban areas) this would not be the case. In the *Results* section, the accuracy achieved by models with varying values of $n$ is considered, with a clear trade-off between efficiency and accuracy evident.

The determination of appropriate starting dimensions *a* x *b*, is a choice individual to each particular map series. It must be chosen to ensure a sufficiently low $R$ value i.e. a high probability that some *a* x *b* areas will contain no features of interest. In the case of the maps examined here, the size was chosen as approximately 16 acres. This was chosen after manual examination of the maps and reflects the general nature of landholdings and settlement patterns in the periods concerned. Fig. 2 shows a tiling from one map sheet (approx.. 1531 m x 665 m = 252 acres) into sixteen approx. 16 acre plots, each measuring 1792 x 768 pixels. The rectangular nature of the tile shape was chosen for convenience, as map sections were obtained from a browser window, accessing the Irish Townland and Historical Map Viewer [20]. While the CNN architectures employed were originally trained on square images (e.g., 224×224), the classifiers still performed well when applied to the fixed-size rectangular inputs used in this study, with fine tuning operations allowing the network to adjust its filters and weights to the new spatial pattern.

Alternative tiling, with larger dimensions can be envisaged by combining a number of the red tiles in Fig 2. e.g. combine four delineated tiles to produce a 3584 x 1536 pixel tile. Using larger alternative initial tile sizes, very few or no areas containing zero features of interest would be identified by the model i.e. high $R$. The initial choice of *a* x *b* = 1792 x 786 pixels was guided by this observation.

A dataset comprising 40 original training images (1792 x 786 pixels; 20 containing buildings and 20 without buildings) was assembled. Images were taken from geographically different map sections to incorporate as much diversity in training data as possible. Data augmentation was applied to each image using horizontal flipping, vertical flipping, 180° rotation, 180° rotation with a horizontal flip, and 180° rotation with a vertical flip, resulting in a total of 240 training images (120 with buildings, 120 without). These transformations all preserved the aspect ratio of the images. A convolutional neural network was trained using the *fastai* vision_learner function with a ResNet18 backbone.

After the initial classification (*n* = 1 stage of the pipeline) of the *a* x *b* sized (1792 x 768 pixel) tiles, further subdivision and re-classification can be conducted iteratively. Subdividing and classifying many times (large *n*) will remove more of the featureless map area, reducing the work required by the segmentation model and improving the efficiency of the pipeline. A factor also contributing to this, is that segmentation models trained on smaller areas will be faster to train and make predictions. However, making *n* too large will result in very small image portions for the segmentation model to address, which may lack sufficient context and features to be accurately segmented, leading to less accurate predictions. Hence, in choosing *n*, there is a tradeoff between model accuracy and speed. In this case, it was decided that *n* = 2 provided the optimal strategy, with each 1792 x 768 pixel tile further subdivided into 21 tiles measuring 256 x 256 pixels. This choice is discussed further in the 'Results' section.

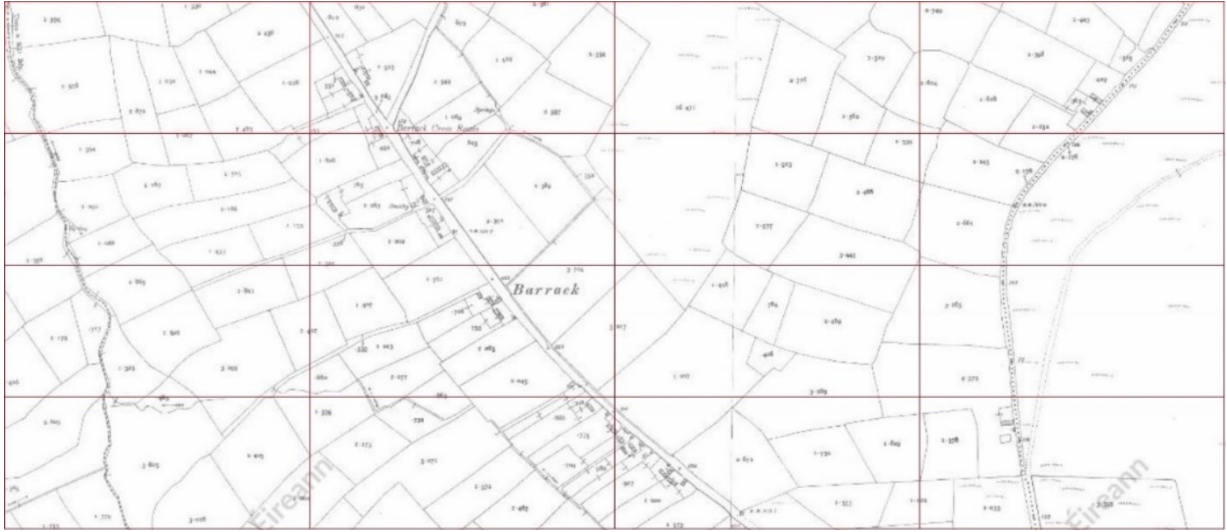

Figure 2: Tiling of large map area (~262 acres) into *a* x b (1792 x 768) pixel areas.

A new classifier was trained on a selection of 256 x 256 tiles. Training an accurate model for these smaller tiles proved more complex than the training of the 1792 x 768 classifier, with the smaller tile size allowing less contextual information for the model. In choosing the training data and parameters, high consideration was given to the desirability of a low number of false negatives, as the pipeline should ideally never discard useful map sections. Some examples of difficult to classify tiles are shown in Fig. 6 and Fig. 9 (see 'Results'). A deliberate class imbalance was maintained in the training data to bias the model towards classifying tiles as containing buildings. The training dataset consisted of 180 'no building' images and 360 'building' images (created by augmenting 60 'building' images using six rotation/flipping operations). A convolutional neural network was again trained using the *fastai* vision_learner function with a ResNet18 backbone.

*Segmentation Model*

The segmentation model was trained on a selection of 50 images, each 256 x 256 pixels in size. These were selected from the same set used to train the *n* = 2 classification model, and were selected to ensure as much variety as possible in the size and shape of buildings sampled. Segmentation masks were created using the Roboflow [24] annotation tools, with smart polygon utilized where possible. This led to slight deviations from straight edges in the mask

(see Fig. 3), but the results were sufficient for the purposes of this study. Masks and images were augmented by flipping horizontally and vertically to generate a training set of 150 images.

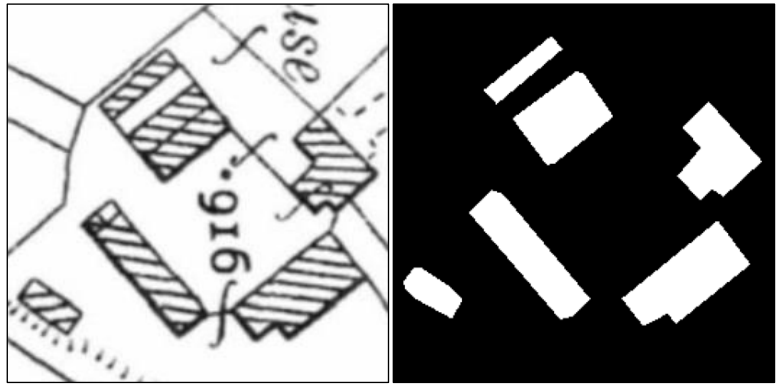

Figure 3: Segmentation model: (left) Training image and (right) training mask produced using Roboflow [24] annotation tools.

A U-Net model with a ResNet-34 backbone was trained using the *fastai* unet_learner for binary semantic segmentation. The model was trained using the BCEWithLogitsLossFlat loss function, to effectively handle the inherent class imbalance between 'building' and 'no building' pixels. A custom Dice metric was implemented for evaluation, as the built-in *fastai* Dice metric does not support raw logits as output by the model. The model was trained for 12 epochs, with validation performed on 30 images. A Dice score of 0.94 was achieved. Fig. 4 shows the model segmentation prediction on two of the validation images. For the purposes of testing the pipeline accuracy (see 'Results'), building centroids were computed as the centre of mass of each connected region of pixels labelled in the binary segmentation mask. To deal with cases where buildings were split across multiple tiles, any tile containing a positive mask section was stitched together with the 8 adjacent tiles. This process was repeated for any tiles in the additional 8 tiles, which had building sections in them. In this way, a perimeter of featureless tiles was built up around the building or building cluster, before applying the centroid determining algorithm. (Fig. 5)

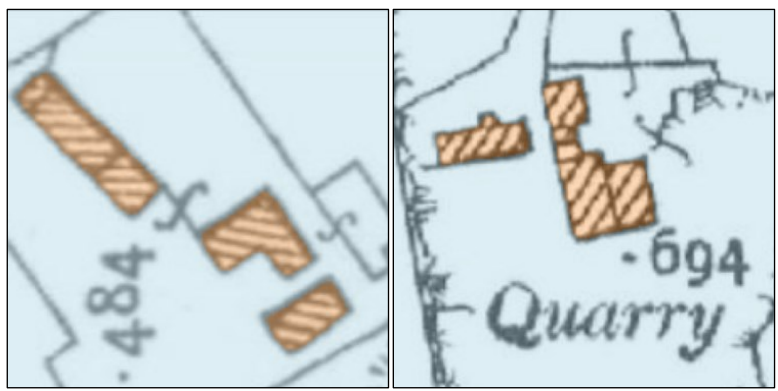

Figure 4: Predicted segmentation of two images from validation set. Pixels identified as 'building' are colored in orange. The mask is overlaid on the original map tile. The difference in quality/resolution of the two map sections is attributable to the fact that they come from different geographical regions.

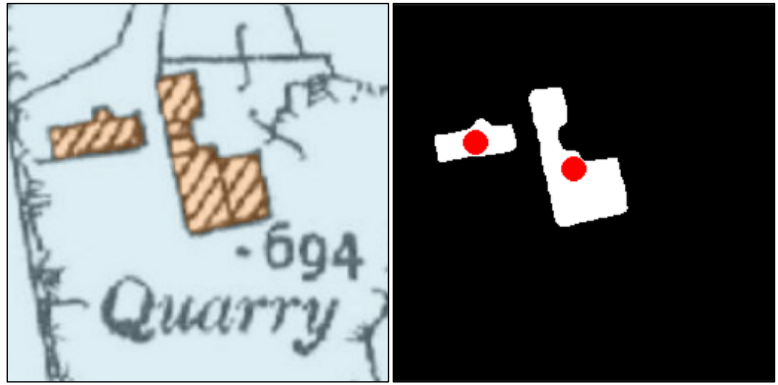

Figure 5: Determining building locations: (left) segmentation superimposed on map tile (right) raw segmentation mask with overlaid building centroids (red dots). Note that this tile was processed with 8 adjacent tiles. Only the central portion is shown.

## Results

*Pipeline Accuracy*

To assess the ability of the pipeline in detecting buildings, three test map areas were selected from the 25 inch map series. As different cartographers of the era used slightly idiosyncratic styles, the areas were selected from three distinct geographic regions in Ireland, one from Co. Cork in the South, one from Co. Galway, in the West, and one from Co. Donegal, in the North. Each map section measured approx. 1531 x 665 m (equivalent to 12 x (1792 x 768 pixel areas)). Each of the three maps was processed through the pipeline as described (using $n$ = 2), the location of buildings found was overlaid on the map, and the map was cross-checked manually to evaluate the accuracy of the method. A total of 36 tiles were examined, within which 54 buildings were manually identified. Of these, 53 were correctly detected by the pipeline, with 1 missed and no false positives recorded, resulting in an F1 score of 0.99. Fig. 6a illustrates a map section where all buildings were accurately identified. Although the $n$ = 1 level classifier achieved an F1 score of 1, inference on some tiles containing single buildings was made with low confidence. An example of such a tile is shown in Fig. 6b. The reduced confidence is presumed to be due to the low resolution of the input image, which degraded the visibility of the cross-hatch pattern, coupled with the small size of the building on the tile. The 'missed' building occurred on tiles mis-classified by the $n$ = 2 classifier (Fig. 6c), and was split across multiple tiles, occurring in the extreme corner or edge of these tiles.

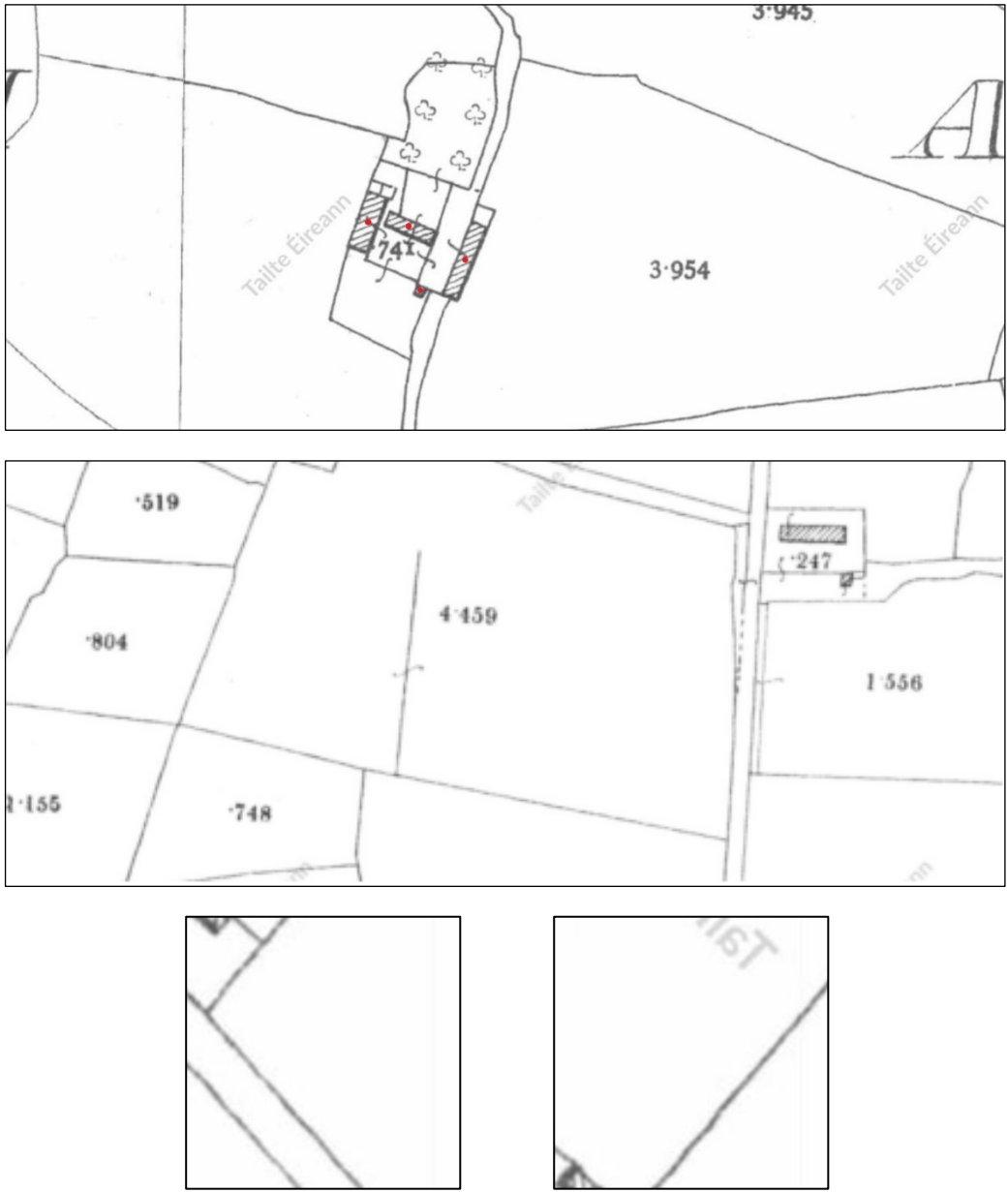

Figure 6: (a) (top) Building centroid (red dots) overlaid on map tiles. These were manually checked to determine model precision. (b) (middle) Tile which was correctly labelled but with a low confidence value (0.5268), by the $n$ = 1 classifier. (c) (bottom) Tiles mislabeled by the $n$ = 2 classifier as 'no building'. The building was small, split across two tiles, and occurred in the extreme edges of those tiles. The mislabeling occurred due to the otherwise largely featureless nature of the tile, and lack of context regarding surroundings.

*Pipeline Efficiency*

As outlined in the methodology section, the gain in computational efficiency of the pipeline depends primarily on the value of *n* (the number of subdivisions or 'tilings' completed). In general, as more classifiers are applied, the area requiring segmentation reduces, and the processing time per acre reduces. However, as classifiers are introduced, the accuracy of the pipeline also reduces. The nature of this trade-off between accuracy and efficiency was investigated by developing pipelines using different values of $n$.

Computational efficiency was estimated using the method outlined in the Methodology section, with values of $R = 0.4$ and $A = 5$ assumed. These were derived from manual observation of map sections in the case of $A$, and timing of classifier and segmentation models applied to large map sections, in the case of $R$. Model accuracy was assessed as outlined in the *Pipeline Accuracy* section. The results are listed in Table 1 and plotted in Fig. 7.

| n | Tile sizes classified | Inference Time Required (normalized) | F1 score |
|---|---|---|---|
| 0 | None | 1 | 1 |
| 1 | 1792 x 768 | 0.6 | 1 |
| 2 | 1792 x 768<br>256 x 256 | 0.44 | 0.99 |
| 3 | 1792 x 768<br>256 x 256<br>128 x 128 | 0.376 | 0.82 |
| 4 | 1792 x 768<br>256 x 256<br>128 x 128<br>64 x 64 | 0.3504 | 0.40 |

**Table 1: Investigation of inference time required for one map area measuring 1531 x 665 m. Inference times have been normalized to 1, to allow for easy comparison.**

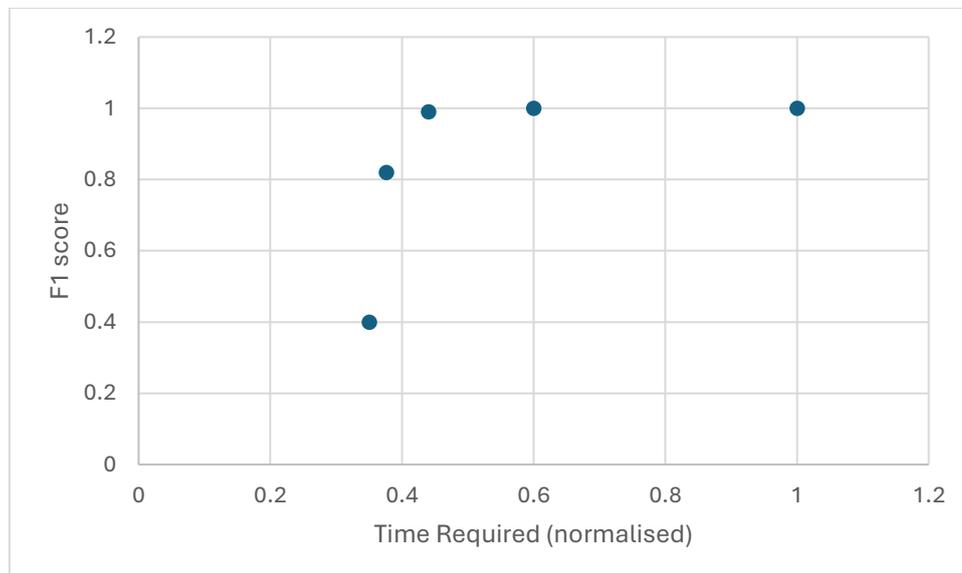

**Figure 7: Plot of F1 score versus computational time required (normalized to 1).**

It is clear that the trade-off between accuracy and efficiency declines above $n = 2$ (normalized time < 0.4). The reason for this can be understood by examining some of the tiles mis-classified by the $n = 2$ classifier (Fig. 6c) and *n* = 3 classifier (Fig. 8) . As the classifier is now processing very small tiles, it frequently processes tiles with only small portions of buildings appearing at the very edges of the tile, and frequently mis-classifies these as containing no buildings. This issue can be considered as a 'lack of context' issue, which is common in classifiers used on images which are too small.

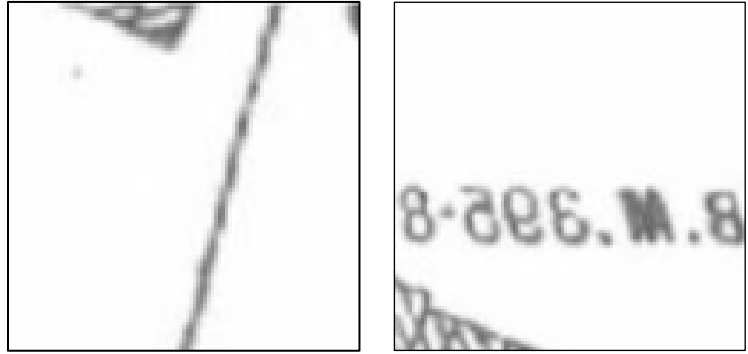

**Figure 8:** *n = 3* **classifier level; 168 x 168 tiles, from the validation set, mislabeled as 'no buildings'. The tiles only containing small portions of buildings were frequently mis-classified.**

The optimum pipeline parameter was set to n=2n = 2n=2. At this stage, most inaccuracies and inefficiencies occur during the final classifier step, which processes 256×256256 \times 256256×256 tiles. This level suffers from a lack of contextual information. Two false negative tiles that reduced the pipeline's accuracy have already been discussed in the 'Pipeline Accuracy' section (Fig. 6c). Additionally, two false positive tiles are illustrated in Fig. 9, where tiles containing no buildings were mistakenly classified as containing buildings. Although the segmentation of these tiles did not result in any false building detections, the misclassification of these tiles as positives introduces inefficiencies in the overall process.

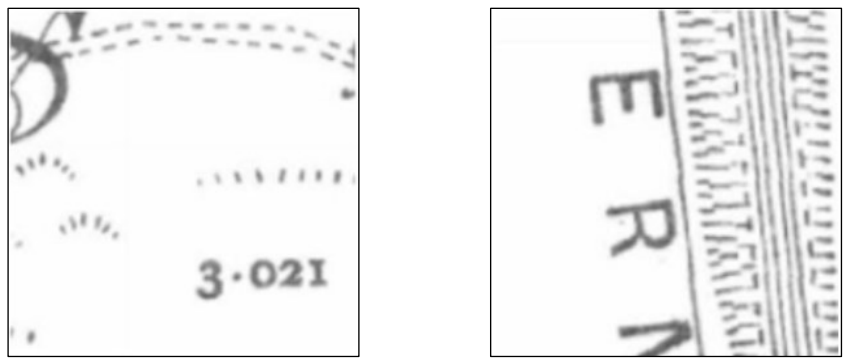

**Figure 9: Tiles mislabeled as containing buildings by the** *n* **= 2 level classifier. (left) the vertical cross-hatching lines which represent wetland are misinterpreted as buildings (right) a feature with dense short line patterns is misinterpreted as a building.**

*Application of Pipeline: Comparison of 6 inch and 25 inch maps*

The same principles described in this paper can clearly be applied to any map series. In addition to the 25 inch map series, Tailte Eireann have also published a 6 inch series map, which is from an earlier time period. The map resolution is comparatively reduced, making the processing of the 6 inch map more difficult than that of the 25 inch map. In addition, although the same notation of cross-hatching is used to identify buildings, individual styles of cartographers was less consistent in the 6 inch series (see Fig. 10) than in the 25 inch series.

To determine if the trained models and pipeline is robust enough to deal with a different map series, the pipeline was applied to the same test areas as described in the 'Pipeline Accuracy' section, using the 6 inch map series. The pipeline was found to be significantly less accurate, yielding an F1 score of 81 %. The errors occurred predominantly in the final segmentation layer. Fig. 10 shows some 6 inch map series 256 x 256 tiles and the associated predicted segmentation masks. A number of false positives are identified. In addition, the generated building shapes are less accurate. It is to be expected that more expansive training of the models for this series, including more varied and extensive training data, and perhaps more high powered models, would improve the overall results. However, even with this reduced level of accuracy, important findings can be made.

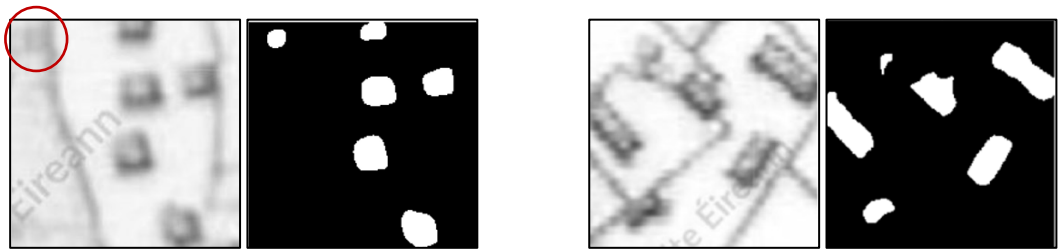

**Figure 10: Segmentation of 256 x 256 tiles (left: original map tiles; right: segmentation masks) from the 6 inch historical series maps. (left) A shaded area which is approximately square shaped on top left of tile is mistaken for a building by the segmentation model (b) Shapes of detected buildings are less clear than in the segmentation model operating on the 25 inch historical map series (see Fig. 4). This is largely due to the lower resolution of the older map series.**

While processing the Co. Galway map test area, which comes from the area of Tully (ITM co-ordinates: 500703, 722113), a comparison of the 6 inch and 25 inch map series identified a settlement (approx. 22 buildings in total) whose buildings had fallen into disrepair in the years between the two surveys. As the two surveys were conducted in 1839 and 1899, these two time points bracket the Great Irish Famine. The pipeline described in this paper is a method of determining areas which lost settlements in that period, and may contain buildings of archaeological interest. Fig. 11 shows the area of Tully on the historical maps and a modern day satellite view on which ruins of the settlement appear to be visible. To my knowledge, this settlement has not previously been identified as a potential 'Famine village', settlements so called because they were largely abandoned due to the effects of the famine. Interestingly, it does not appear on the 1841 census report [25] listed as a town (defined as an 'agglomeration' which contained 20 or more houses), and consequently does not appear on the list of 'towns' which disappeared between the 1841 and 1851 censuses assembled by Crowley et al. [26].

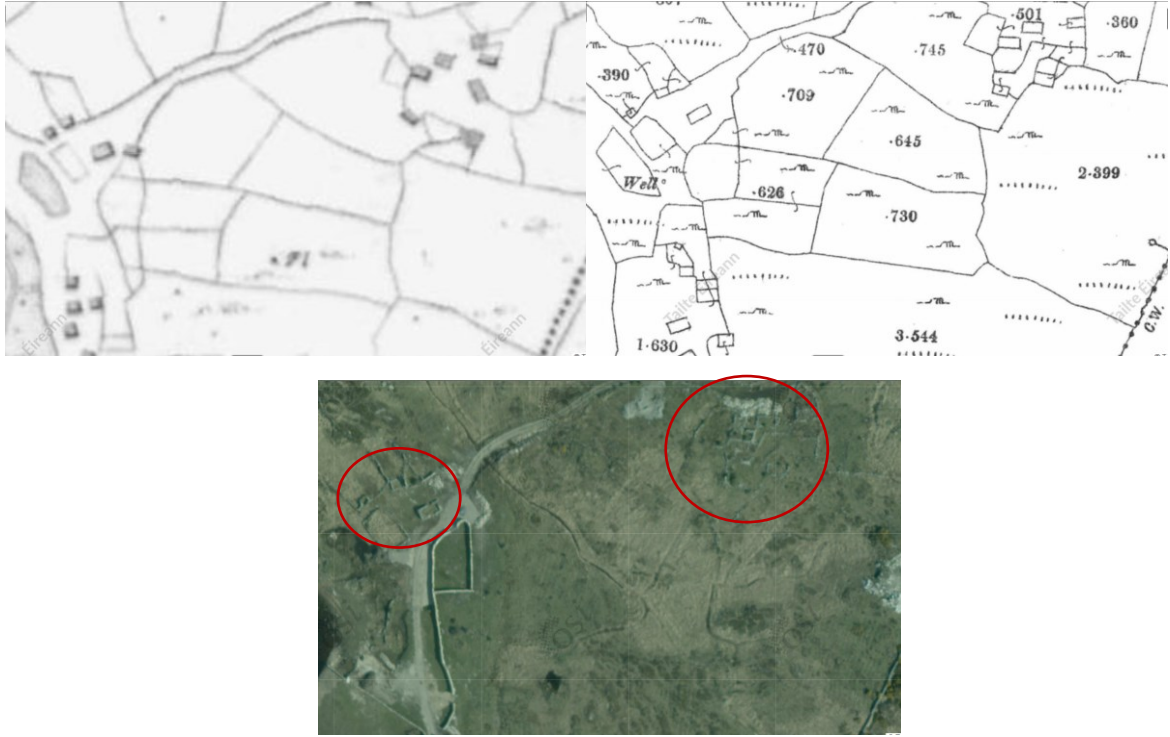

Figure 11: Tully, Co. Galway (ITM co-ordinates: 500703, 722113). (top left) 6 inch map series (1839; approx. 6 years before the Great Famine) 14 buildings are visible. (top right) 25 inch map series (1899; approx. 50 years after the Great Famine), where buildings from 1839 map no longer appear. (bottom) Modern day image. Possible sites of archaeological interest highlighted in red.

**Conclusion**

A novel pipeline for building footprint detection in historical rural maps has been developed, offering improved efficiency over traditional full-map segmentation methods. It uses an iterative classification process to exclude non-building areas early in the workflow, significantly optimizing processing.

Tested on Irish historical maps, the pipeline demonstrated high accuracy, though precision was somewhat reduced on the older, lower-resolution 6-inch map series due to occasional mislabelling.

The method also shows promise for identifying 'lost' or 'vanished' settlements, which may of historical and/or archaeological interest. Application of the pipeline to two map series from different time periods revealed a settlement consisting of 22 buildings in Tully, Co. Galway, visible in the 1839 6-inch pre-Famine map, but absent from the 1899 25-inch post-Famine map.

Future work will aim to further improve the performance of the pipeline. This may be achieved by enhancing and diversifying the training dataset, or by incorporating an additional border around smaller tiles—an approach used by [2] —to improve classifier accuracy. However, this method comes with the drawback of increased computational cost. To mitigate this, the integration of a more efficient building footprint extraction algorithm, such as the one proposed by [4], will be considered. Another key objective will be the development of a user-friendly interface for accessing and applying the method, with the goal of making it more accessible to non-specialist users, including historians and researchers in the humanities.